\documentclass[conference]{IEEEtran}
\IEEEoverridecommandlockouts
\usepackage{cite}
\usepackage{amsmath,amssymb,amsfonts}
\usepackage{algorithmic}
\usepackage{graphicx}
\usepackage{textcomp}
\usepackage{xcolor}
\usepackage[hidelinks]{hyperref}
\usepackage{booktabs}

\usepackage{soul}

\def\BibTeX{{\rm B\kern-.05em{\sc i\kern-.025em b}\kern-.08em
    T\kern-.1667em\lower.7ex\hbox{E}\kern-.125emX}}
\begin{document}

\title{Robust Autonomous UAV Landing on Maritime Platforms via Multimodal Agentic AI and Active Wave Compensation
\thanks{$^*$Authors have contributed equally to this work. This project has received funding from the European Union’s Horizon Europe research and innovation programme under Grant Agreement No. 101189723 (AEROSUB); The study was supported by Fundação Para a Ciência e a Tecnologia (FN: 2024.01361.BD and PP: 2022.11442.BD (\url{doi.org/10.54499/2022.11442.BD}).}
}

\author{\IEEEauthorblockN{
    Francisco S. Neves*,
    Pedro N. Pereira*, Raul D.S.G. Campilho\\
    Andry M. Pinto
    }
    
    \IEEEauthorblockA{
    \emph{Faculty of Engineering, University of Porto (FEUP)}\\
    \emph{Centre for Robotics and Autonomous Systems - INESC TEC} \\
    \emph{Instituto Superior de Engenharia do Porto (ISEP)} \\
    Porto, Portugal
    \\\{francisco.s.pinto, pedro.n.almeida\}@inesctec.pt}
    }

\maketitle

\begin{abstract}
Autonomous aerial inspection of marine infrastructure is frequently compromised by stochastic sea states, introducing risks of high-kinetic impacts, post-landing toppling, and sensory occlusion. This paper proposes a decoupled, multi-vehicle landing framework synchronizing an Unmanned Surface Vehicle (USV) equipped with a 3-RPU stabilized platform with a robust Unmanned Aerial Vehicle (UAV). The architecture utilizes two independent Deep Reinforcement Learning (DRL) agents: a Soft Actor-Critic (SAC) agent providing high-frequency wave-motion compensation for the landing deck, and a multimodal RL agent for the UAV’s final approach. Evaluated in high-fidelity maritime simulations, the system achieved a 100\% landing success rate across 15 trials in wave states varying from calm to rough. Results show a mean stabilization efficacy of 87.8\%, maintaining the landing surface within 1° of the horizontal plane for 96\% of the mission duration in rough conditions, effectively contributing to safer landings.
\end{abstract}

\begin{IEEEkeywords}
Robotics, Autonomous UAV Landing, USV Wave Compensation
\end{IEEEkeywords}
\section{Introduction}
The inspection of maritime infrastructures, such as offshore wind turbines and oil rigs, remains a high-stakes task that often places human personnel in hazardous environments. Autonomous aerial robots offer a viable path toward safer and more objective inspections, yet their deployment is frequently limited by the energy density of flight batteries and the extreme volatility of the maritime landing environment \cite{diuus, pinto2021atlantis}. To overcome these constraints, this paper presents a heterogeneous multi-vehicle system where an Uncrewed Surface Vehicle (USV) and an Uncrewed Aerial Vehicle (UAV) perform independent but complementary roles within a single mission.

To prevent the long-range transit between the shore and the infrastructure location, the USV can serve as a flexible autonomous landing platform. The primary challenge in such a system is the landing phase, where wave-induced roll and pitch motions create a highly unstable target \cite{machines11040478}. Traditional landing strategies often place the entire burden of compensation on the UAV, a task that becomes affected in GNSS-denied or sensory-degraded conditions \cite{claro2024offshore}. To address these challenges without the need for cross-platform communication, this paper introduces a complementary stabilization strategy. This approach splits the operational burden between the two platforms to ensure a robust landing interface. On the USV, a 3-Degree-of-Freedom (3-DoF) parallel mechanism. Reactive stabilization systems like PID, or SMC often suffer from "chasing behaviors" in dynamic sea states, failing to maintain the level surface necessary for a safe touchdown \cite{6932761, 10704910}. Instead, our approach resorts to Deep Reinforcement Learning (DRL) to independently mitigate wave motion, providing a sub-degree stabilized landing platform. The landing pad stabilization is further complemented by a simple PID position hold, that benefits from an omnidirectional "cross" configuration. On the other hand, in UAV navigation, traditional perception-action pipelines are typically built on sequential modules that can fail entirely if a single feature is occluded \cite{lit_pipeline0, lit_pipeline1}. Thus, we employ a multimodal Transformer-based agentic architecture to fuse visual, thermal, and LiDAR data, ensuring resilient navigation through sensory uncertainty as proposed in \cite{neves2024multimodal}. By combining a dynamically stabilized landing target with an agentic aerial landing policy, the system demonstrates a successful landing through the complementary robustness of each vehicle's strategy.

The key contributions of this work are:
\begin{enumerate}
\item \textbf{Reliable Multi-Vehicle Landing:} A mission profile that maximizes UAV inspection time by utilizing an independent USV hub for logistics and energy management, ensuring that landing occurs in a stabilized zone away from hazardous structures.
\item \textbf{Independent Wave Mitigation via DRL:} A 3-RPU (Revolut Prismatic Universal) mechanism controlled by SAC (Soft Actor-Critic) that achieves an 86.1\% motion reduction to facilitate touchdown without requiring input from the aerial vehicle.
\item \textbf{Resilient Agentic UAV Landing:} A multimodal agentic policy for the UAV that maintains persistent navigation under diverse sensor degradations, achieving a field-proven precision of 0.10±0.08 m without requiring external positioning systems.
\end{enumerate}

\section{Complementary Multi-Vehicle Landing}

\begin{figure}
\label{fig:task}
\centering
\includegraphics[width=0.75\columnwidth, trim=0pt 40pt 0pt 0pt, clip]{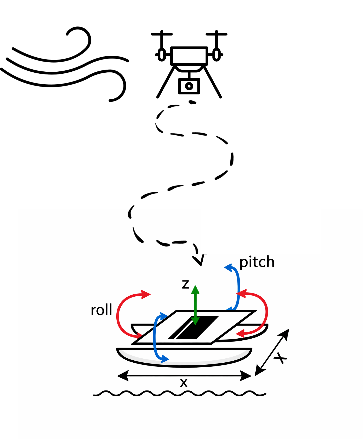}
\caption{Representation of the multi-vehicle landing task. The UAV and ASV are disturbed by wind and wave oscillations. }
\end{figure}

The task of autonomous multi-vehicle landing, depicted in \ref{fig:task}, can suffer from multiple issues, especially when considering highly active sea states with gusts of wind that can easily destabilize both the UAV and USV. For this task we've identified three catastrophic scenarios:
\begin{itemize}
    \item \textbf{1. High-energy kinetic impact:} In uncompensated scenarios, the phase mismatch between the drone's descent velocity and the platform's vertical heave or angular tilt-rate can result in a high-velocity collision. Even if the drone's RL policy is robust, a sudden upward collision from the deck during the final throttle-down phase can cause a tip-over (toppling). 

    \item \textbf{2. Static instability and toppling:} Landing on an uncompensated, inclined surface shifts the drone's Center of Mass (CoM) relative to its footprint. Even a successful touchdown can result in a post-landing topple if the deck's tilt exceeds the drone's stability margin or if the landing gear's friction is insufficient to counteract the lateral gravitational component. 
    
    \item \textbf{3. Translational escape during the final meter disconnect):} During the critical final descent phase ($<1$~m), the drone's downward-facing camera experiences a significantly narrowed Field of View (FoV). In an uncompensated system rapid lateral displacements may occur, which can sweep the Aruco marker or thermal docking targets out of the sensor frame faster than the drone's navigation stack can react.
\end{itemize}

Our proposed framework aims to address and negate each of these scenarios by providing a multi-vehicle strategy that operates without any active cross-platform communication. The surface vehicle focuses on providing a stabilized autonomous platform, while the aerial vehicle independently manages the final approach through its own internal multimodal perception. By decoupling the landing deck's mechanical stabilization from the drone's navigation logic, the system remains robust against both physical wave disturbances and sensory uncertainty. 

The proposed architecture splits the operational complexity of maritime landing into three independent, complementary control layers:

\paragraph{USV stabilizer for wave-mitigation:}
The control logic for the wave-motion stabilizer is a model-free Deep Reinforcement Learning (DRL) agent trained using the Soft Actor-Critic (SAC) algorithm. This controller learns the non-linear mapping between vessel IMU data and platform displacement, which is interpreted into actuator displacement via inverse kinematics and handled with off-the-shelf drivers that maintain a precision of $\pm1$ RPM, actively counteracting wave disturbances to maintain a level landing deck. It also benefits from a relatively low size with a state input vector of 10 components: vessel roll/pitch, vessel roll/pitch rates, platform roll/pitch, platform roll/pitch rates, and platform roll/pitch errors. This input passes through a lightweight network:
\begin{equation*}
    \mathcal{N}128\rightarrow\mathcal{N}128\rightarrow\mathcal{N}64\rightarrow\mathcal{N}64\rightarrow[\dot{a_{roll}},\dot{a_{pitch}}],
\end{equation*}
outputting continuous actions for pitch and roll rate commands for the top platform. To complement the RL-based orientation control, a secondary regulatory layer handles heave stabilization via a PID controller. This controller targets the vertical linear velocity derived from the vessel's IMU. To mitigate the inherent integration bias associated with low-cost inertial sensors, a high-pass filter is applied to continuously drive the velocity setpoint toward zero. Furthermore, a centering term is implemented to bias the 3-RPU actuators toward their optimal orientation workspace position.
By maintaining a near-zero relative angular velocity, the stabilizer ensures a more robust approach to critical scenarios 1 and 2, such that the kinetic energy at touchdown is governed mainly by the drone's controlled descent rather than the ship's stochastic motion, and by ensuring a near-level landing surface, we effectively neutralize the risk of the drone sliding into deck infrastructure or tipping overboard.
\hfill

\paragraph{USV stabilizer for position hold:}
The position hold in $x$, $y$, and $yaw$ is achieved through a cascaded PID (position $\rightarrow$ velocity $\rightarrow$ thrust) using GPS pose estimation to softly anchor the USV. To mitigate the stochastic nature of wave-induced roll and pitch, the USV's 3-RPU (Revolute-Prismatic-Universal) parallel mechanism. This addresses the third critical case, which ensures that the platform remains visible to UAV perception system. While the GPS-based hold position is subject to coarse global drift and multi-path interference near large maritime structures, its primary role is to maintain the landing platform within the UAV's operational sensing capabilities.
\hfill

\begin{figure}
\includegraphics[width=\columnwidth]{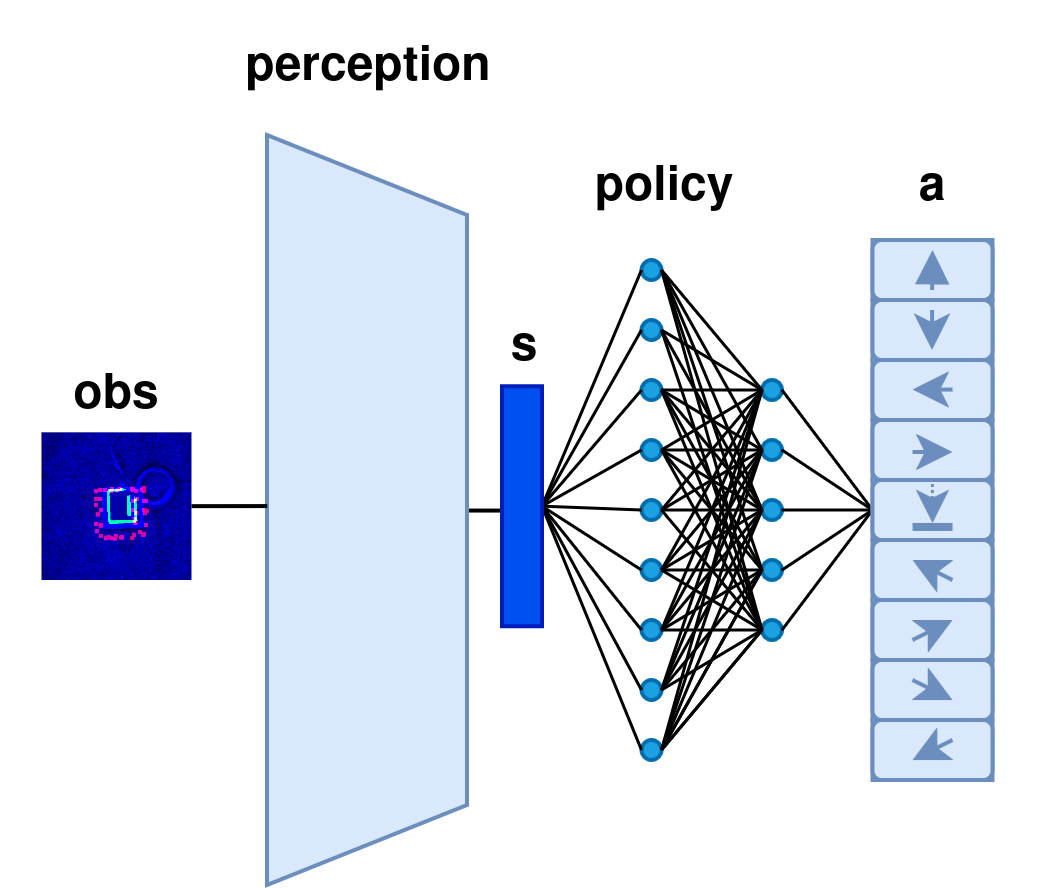}
\caption{Raw sensory data ($obs$) is processed by the multimodal perception module into a latent state ($s$) representation. This state serves as input for the policy, which generates the navigation action ($a$) to be executed.}
\label{fig:agent}
\end{figure}

\paragraph{UAV Agentic Landing}
The UAV uses a behavioral-level agentic formulation for autonomous landing, following the approach proposed in \cite{neves2024multimodal}. A multimodal learning-based approach for autonomous landing of UAV. Instead of relying on a conventional perception–state estimation–control pipeline, the method maps raw multimodal inputs (visual, thermal, and LiDAR) directly to high level navigation actions. The architecture combines perception and decision-making within a single model, where a multimodal perception module, based on a Transformer-like detector, fuses the different sensor inputs into a learned latent representation. This representation is then processed by a policy network trained with reinforcement learning to output discrete navigation actions during the approach and descent phases. By avoiding explicit pose estimation and intermediate geometric representations, the method focuses on learning action-oriented behavior from observations while still interfacing with a standard low-level flight controller. This design allows the system to operate independently of specific platform dynamics and maintain stable guidance even under degraded sensing or GNSS-denied conditions. The agentic policy is illustrated in Figure \ref{fig:agent}.
\hfill

\section{Results}
This section presents a quantitative evaluation of the multi-vehicle system's performance under diverse maritime conditions. The objective is to validate the reliability of the decoupled architecture, demonstrating how independent platform stabilization facilitates a safer and more efficient landing process for the agentic UAV without requiring cross-platform communication. 

In Section \ref{subsec:setup}, it is described the experimental configuration, including the simulated sea states and sensory hardware. In Section \ref{subsec:landing_stab_results}, it is presented the landing and stabilization results.

\subsection{Experimental Setup}\label{subsec:setup}
The proposed UAV-ASV system is evaluated in a Gazebo-based simulation environment configured for realistic maritime conditions. The setup includes:
\begin{itemize}
    \item A quadrotor UAV (IRIS model) equipped with a gimbal-mounted sensor suite—comprising a visual camera, a thermal camera, and a 3D LiDAR—providing the multimodal raw data required for the agentic navigation policy.
    
    \item The USV is equipped with a 3-RPU landing platform (Nest USV) and simulated in a dynamic maritime environment. Wave fields are synthesized using Tessendorf’s FFT method based on the ECKV spectrum. To ensure high-fidelity motion, the simulation employs a mesh-based hydrostatic model rather than simplified point-buoyancy. This allows for real-time calculation of instantaneous buoyancy and restoring moments by computing the submerged hull volume as it intersects with the evolving ocean surface.
        
    \item An ArTuga landing target\cite{claro2023artuga}. It is an artificial multimodal fiducial marker mounted at the center of the ASV platform.
\end{itemize}

\begin{figure}
    \centering
    \includegraphics[width=1\linewidth]{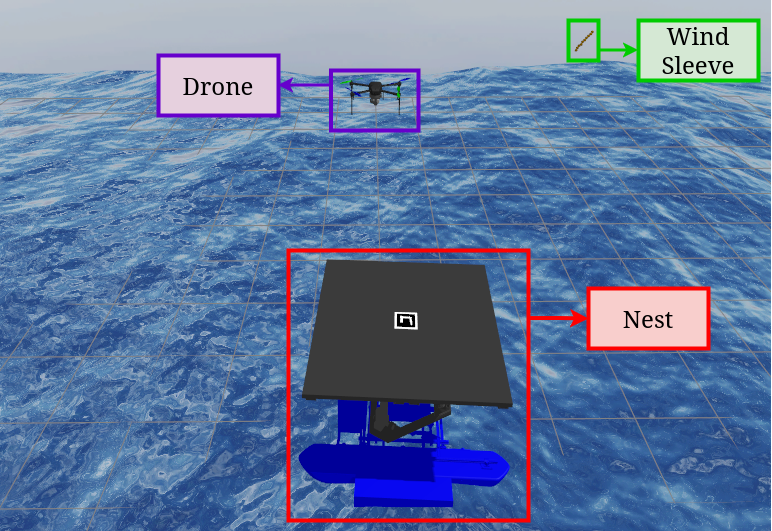}
    \caption{Gazebo Harmonic wave-field simulation with a Quadrotor UAV, the Nest ASV, and a wind-sleeve indicator.}
    \label{fig:simulator}
\end{figure}

To assess the robustness of the decoupled architecture, the simulation environment introduces stochastic wave-induced disturbances that trigger the USV's independent 6-DoF stabilization. The UAV’s agentic policy produces navigational twist commands (linear and angular velocities) based on its onboard perception, while low-level stabilization is delegated to the internal flight controller. The integrated deployment and testing were conducted on a workstation with an Intel® Core™ i5-10600K CPU, NVIDIA GeForce RTX 3060 GPU, 32 GiB RAM, and MS-7C73 motherboard running Ubuntu 24. The simulator interface, showcasing the UAV and the stabilized landing platform, is depicted in Figure \ref{fig:simulator}.

\subsection{Landing and Stabilization}\label{subsec:landing_stab_results}

The proposed system was evaluated in a high-fidelity maritime simulation environment across 15 landing trials under three different sea states: calm, moderate, and rough. 

\begin{table*}
\centering
\caption{Autonomous Landing Performance Across Varying Sea States. Calm: Trials 1-5; Moderate: Trials 6-10; Rough: Trials 11-15.}
\label{tab:landing_results}
\resizebox{2\columnwidth}{!}{%
\begin{tabular}{ccccccccc}
\toprule
Trial & Wind speed (m/s) & Wind Direction ($^\circ$) & Mean Platform Tilt ($^\circ$) &  Mean Nest Tilt ($^\circ$) & Efficacy (\%) & Below 1$^\circ$ (\%) & Center Distance (m)  &  Time (s) \\ 
\midrule
1 & 2.0 & 0.0 & 0.599 & 0.087 & 85.51 & 100.00 & 0.101 & 21.81 \\
2 & 2.0 & 22.5 & 0.678 & 0.095 & 86.02 & 100.00 & 0.050 & 22.33 \\
3 & 2.0 & 45.0 & 0.625 & 0.095 & 84.77 & 100.00 & 0.090 & 26.00 \\
4 & 2.0 & 67.5 & 0.586 & 0.086 & 85.39 & 100.00 & 0.171 & 24.55 \\
5 & 2.0 & 90.0 & 0.615 & 0.089 & 85.57 & 100.00 & 0.066 & 20.34 \\ \midrule
6 & 4.0 & 0.0 & 2.364 & 0.217 & 90.83 & 100.00 & 0.025 & 36.02 \\
7 & 4.0 & 22.5 & 2.038 & 0.297 & 85.42 & 99.06 & 0.572 & 59.91 \\
8 & 4.0 & 45.0 & 2.052 & 0.284 & 86.15 & 98.23 & 0.050 & 47.36 \\
9 & 4.0 & 67.5 & 1.978 & 0.201 & 89.85 & 99.56 & 0.626 & 54.77 \\
10 & 4.0 & 90.0 & 2.131 & 0.203 & 90.45 & 100.00 & 0.574 & 51.20 \\ \midrule
11 & 6.0 & 0.0 & 2.952 & 0.303 & 89.72 & 97.12 & 0.357 & 83.71 \\
12 & 6.0 & 22.5 & 2.928 & 0.304 & 89.62 & 98.23 & 0.391 & 97.24 \\
13 & 6.0 & 45.0 & 3.060 & 0.300 & 90.18 & 97.98 & 0.101 & 55.86 \\
14 & 6.0 & 67.5 & 2.883 & 0.331 & 88.50 & 96.12 & 0.406 & 41.89 \\
15 & 6.0 & 90.0 & 3.016 & 0.290 & 90.39 & 98.56 & 0.577 & 83.11 \\ \bottomrule
\end{tabular}%
}
\end{table*}

The results in Table \ref{tab:landing_results} demonstrate high operational reliability across all sea states. The 3-RPU platform achieved a mean stabilization efficacy of 87.9\%. While the vessel experienced tilts up to 3.06$^\circ$ (Trial 13), the Nest maintained the landing surface below a mean tilt of 0.33$^\circ$. Critically, the surface remained within 1$^\circ$ of the horizontal plane for over 96\% of the mission duration in the roughest conditions (WS 6.0 m/s), neutralizing risks of high-kinetic impact and post-landing toppling.

The system achieved a 100\% landing success rate. In calm and moderate states (Trials 1--10), mean landing error remained under 0.6 m. Mission time increased in rough states, peaking at 97.24 s (Trial 12), as the UAV’s RL policy prioritized safety by awaiting stabilized landing windows. While the GPS-based position hold partially mitigated "translational escape," its algorithmic simplicity proved to be a limiting factor in harsher wave states.

Ultimately, the data suggests a decoupling of mission success from environmental severity. In uncompensated scenarios, these sea states typically induce sensory occlusion or mechanical instability, however, our dual-vehicle approach ensures consistent target perception. Although GPS drift increased center-distance errors in Trials 7, 9, and 15, the level deck provided by the Nest ensured these landings remained statically stable and safe for the UAV.

\section{Conclusions}
This paper presented a decoupled, dual-agent framework for autonomous aerial landing on a maritime vessel under active sea states. By integrating an RL-based wave-motion stabilizer with a multimodal RL landing policy, the system effectively mitigates the risks of high-kinetic impact, post-landing toppling, and sensory occlusion. Experimental results across 15 high-fidelity simulation trials demonstrate a 100\% landing success rate and an average stabilization efficacy of 87.8\%, maintaining the landing deck within 1$^\circ$ of the horizontal plane for 96\% of the mission duration in rough seas. These results suggest that mechanical deck stabilization and robust landing target detection is a critical enabler for safe, objective maritime inspections in hazardous environments.

Future research will focus on: 
\begin{itemize}
    \item Transitioning from high-fidelity simulation to real-world maritime deployment to validate the system's performance under atmospheric disturbances and varying lighting conditions. 
    \item Replacing the current GPS-based hold position with a more robust, sensor-fused localization suite to reduce translational drift.
    \item Implementing a bi-directional communication link between the UAV and USV to share state estimates and synchronize control actions.
\end{itemize}

\bibliography{myrefs}
\bibliographystyle{ieeetr}

\vspace{12pt}
\color{red}

\end{document}